\documentclass[journal]{IEEEtran}

\usepackage[fleqn]{amsmath}
\usepackage{algorithmic}
\usepackage{array}
\usepackage{amssymb}
\usepackage{graphicx}
\usepackage{threeparttable}
\usepackage{multirow}
\usepackage{colortbl}
\usepackage{caption}
\usepackage{subfig}
\usepackage{url}
\usepackage{bm}
\usepackage{rotating}
\usepackage{multirow}
\usepackage{latexsym}
\usepackage{cite}

\begin{document}
\title{Sleep Stage Classification Based on Multi-level Feature Learning and Recurrent Neural Networks via Wearable Device}

\author{Xin Zhang, Weixuan Kou, Eric I-Chao Chang, He Gao, Yubo Fan and Yan Xu*
\thanks{This work is supported by Microsoft Research under the eHealth program, the National Natural Science Foundation in China under Grant 81771910, the National Science and Technology Major Project of the Ministry of Science and Technology in China under Grant 2017YFC0110903, the Beijing Natural Science Foundation in China under Grant 4152033, the Technology and Innovation Commission of Shenzhen in China under Grant shenfagai2016-627, the Beijing Young Talent Project in China, the Fundamental Research Funds for the Central Universities of China under Grant SKLSDE-2017ZX-08 from the State Key Laboratory of Software Development Environment in Beihang University in China, the 111 Project in China under Grant B13003. \emph{* indicates corresponding author.}}
\thanks{Xin Zhang, Weixuan Kou, Yubo Fan and Yan Xu are with the State Key Laboratory of Software Development Environment and the Key Laboratory of Biomechanics and Mechanobiology of Ministry of Education and Research Institute of Beihang University in Shenzhen and Beijing Advanced Innovation Centre for Biomedical Engineering, Beihang University, Beijing 100191, China (email: xinzhang0376@gmail.com; weixuankou@outlook.com; yubofan@buaa.edu.cn; xuyan04@gmail.com).}
\thanks{Eric I-Chao Chang, and Yan Xu are with Microsoft Research, Beijing 100080, China (email:echang@microsoft.com; xuyan04@gmail.com).}
\thanks{He Gao is with Clinical Sleep Medicine Center, the General Hospital of the Air Force, Beijing 100142, China (email:bjgaohe@sohu.com).}
}

\maketitle

\begin{abstract}
This paper proposes a practical approach for automatic sleep stage classification based on a multi-level feature learning framework and Recurrent Neural Network (RNN) classifier using heart rate and wrist actigraphy derived from a wearable device. The feature learning framework is designed to extract low- and mid-level features. Low-level features capture temporal and frequency domain properties and mid-level features learn compositions and structural information of signals. Since sleep staging is a sequential problem with long-term dependencies, we take advantage of RNNs with Bidirectional Long Short-Term Memory (BLSTM) architectures for sequence data learning. To simulate the actual situation of daily sleep, experiments are conducted with a resting group in which sleep is recorded in resting state, and a comprehensive group in which both resting sleep and non-resting sleep are included. We evaluate the algorithm based on an eight-fold cross validation to classify five sleep stages (W, N1, N2, N3, and REM). The proposed algorithm achieves weighted precision, recall and $F_{1}$ score of 58.0\%, 60.3\%, and 58.2\% in the resting group and 58.5\%, 61.1\%, and 58.5\% in the comprehensive group, respectively. Various comparison experiments demonstrate the effectiveness of feature learning and BLSTM. We further explore the influence of depth and width of RNNs on performance. Our method is specially proposed for wearable devices and is expected to be applicable for long-term sleep monitoring at home. Without using too much prior domain knowledge, our method has the potential to generalize sleep disorder detection.
\end{abstract}

\begin{IEEEkeywords}
Heart rate, Long Short-Term Memory, Recurrent neural networks, Sleep stage classification, Wearable device.
\end{IEEEkeywords}

\section{Introduction}
\IEEEPARstart{S}{LEEP} is a fundamental physiological activity of the human body, which contributes to self-recovery and memory consolidation \cite{stickgold2005sleep,carskadon2005normal}. Regular sleep facilitates the performance of daily work. However, many sleep disorders, such as insomnia, apnea, and narcolepsy, disturb sleep quality and thus threaten human health \cite{american2005international}. Effective diagnosis and treatment of these sleep disturbances rely on accurate detection of sleep and sleep cycles \cite{li2017hyclasss}. Therefore, sleep stage classification is a premise and significant step for sleep analysis.

The standard technique for scoring sleep is to use polysomnography (PSG) to synchronously record multichannel biological signals which include electroencephalogram (EEG), electrooculogram (EOG), electromyogram (EMG), electrocardiogram (ECG), respiratory effort signals, blood oxygen saturation, and other measurements all through the night in a hospital. These recordings are divided into nonoverlapping 30-second epochs. Domain experts evaluate sleep epoch by epoch, based on Rechtschaffen and Kales (R\&K) rules \cite{rechtschaffen1968manual} and the more recent American Academy of Sleep Medicine (AASM) \cite{berry2012aasm} guideline. According to the AASM, sleep is categorized into five stages: Wake (W), Rapid Eye Movement (REM) and Non-Rapid Eye Movement (NREM, including N1, N2, and N3).

Monitoring sleep through the PSG system has many disadvantages with respect to home use. First, patients have to wear numerous sensors on different parts of the body. It may negatively impact patients' normal sleep and thus produce discrepant results which are not able to reflect the actual sleep condition. Second, PSG is expensive, which makes it not available for most ordinary families. Third, PSG is not portable, making it inappropriate for long-term home monitoring. To overcome the above shortcomings, utilizing a wearable device in place of the PSG system to classify sleep stages automatically is a promising strategy. A wearable device can readily record heart rate and body movement without causing many obstructions to natural sleep. Additionally, relationships between sleep stage transition and both heart rate and body movement have been extensively investigated in previous studies \cite{task1996heart,cole1992automatic}. By designing a suitable algorithm based on such a principle, the wearable device can be an alternative choice to score sleep automatically at home.

Many physiological studies of sleep have indicated that sleep structure is associated with autonomic nervous system (ANS) regulation \cite{baharav1995fluctuations,trinder2001autonomic,task1996heart}. The contribution of parasympathetic and sympathetic activity varies between different sleep stages. Meanwhile, ANS activity during sleep can be measured using heart rate variability (HRV) as a quantitative index of parasympathetic or sympathetic output \cite{otzenberger1998dynamic,jerath2014role,bonnet1997heart}. Hence, it is reasonable to consider making use of heart rate for determining sleep stages. More specifically, the sympathetic input is reduced and parasympathetic activity predominates in NREM sleep. Thus heart rate decreases, with less variability. During REM sleep, in contrast, sympathetic activity shows more predominate influence. Accordingly, heart rate increases and becomes more unstable \cite{nazeran2006heart}. The spectral components of heart rate also exhibit different characteristics in sleep transition \cite{malik1996task,baharav1995fluctuations}. The ratio of the power in low frequency (LF, 0.04-0.15 Hz) to high frequency (HF, 0.15-0.4 Hz) tends to decrease in NREM sleep and significantly increase in REM sleep.

Quite a few sleep staging methods have been developed based on HRV, with most of them emphasizing the applicability in home-based scenarios. Yoon et al. \cite{yoon2017slow} designed thresholds and a heuristic rule based on automatic activations derived from HRV to determine the slow wave sleep (SWS). An overall accuracy of 89.97\% was achieved. Long et al. \cite{long2017detection} also designed an SWS detection method using cardiorespiratory signals. With a feature smoothing and time delay procedures, they achieved a Cohen's Kappa coefficient of 0.57. Ebrahimi et al. \cite{ebrahimi2013automatic} extracted time-domain features, nonlinear-dynamics features and time-frequency features from HRV using empirical mode decomposition (EMD) and discrete wavelet transform (DWT), followed by linear discriminant (LD) and quadratic discriminant (QD) classifiers. Xiao et al. \cite{xiao2013sleep} extracted 41 HRV features in a similar way and random forest (RF) \cite{breiman2001random} was used to classify 3 different sleep stages (wake, REM, and NREM). A mean accuracy of 72.58\% was achieved in the subject independent scheme.

Actigraphy-based methods which capture body movement during sleep have long been investigated, especially to identify wake/sleep \cite{cole1992automatic,sadeh1994activity-based,paquet2007wake}. They are easy to understand since the body tends to remain stationary when we fall asleep, and the motion amplitude becomes distinctively smaller than in wake state. Various studies have been implemented to evaluate sleep staging with actigraphy. Herscovici et al. \cite{herscovici2007detecting} presented an REM sleep detection algorithm based on the peripheral arterial tone (PAT) signal and actigraphy which were recorded with an ambulatory wrist-worn device. Kawamoto et al. \cite{kawamoto2013actigraphic} demonstrated that respiratory information could be estimated from actigraphy and detected REM sleep based on detected respiratory rate. Long et al. \cite{long2014sleep} designed features based on dynamic warping (DW) methods to classify sleep and wake using actigraphy and respiratory efforts.

So far, few sleep staging methods have been developed that use both heart rate and wrist actigraphy. Note that the two signals can be easily derived together by a wearable device, it is worth attempting to design such a kind of algorithm. Furthermore, most protocols in previous studies focus on designing low-level features which are extracted in the time domain, frequency domain, and nonlinear analysis. This causes the effectiveness of feature extraction to be overly dependent on the expertise analysis of signals, which makes these hand-engineered features not robust and flexible enough to adapt different circumstances. In this paper, we propose a multi-level feature learning framework which extracts low- and mid-level features. Low-level features capture temporal and frequency domain properties and mid-level features learn compositions and structural information of signals. Specifically, the mean value and Discrete Cosine Transform (DCT) \cite{quiceno2009detection} are adopted to heart rate and cepstral analysis \cite{furui1981cepstral} is adopted to wrist actigraphy to extract low-level features, respectively. Mid-level features are extracted based on low-level ones.

Recently, deep learning theories have made dramatic progress, based on which state-of-the-art results are achieved in many important applications. With complicated network architectures, deep learning methods can mine information without finely designed features, showing stronger robustness. Tsinalis et al. \cite{tsinalis2016automatic} utilized convolutional neural networks (CNNs) based on single-channel raw EEG to predict sleep stages without using prior domain knowledge. The sparse deep belief net (DBN) was applied in \cite{zhang2016automatic} as an unsupervised method to extract features from EEG, EOG and EMG. Sleep staging is a sequential problem \cite{berry2012aasm} since sleep shows typically cyclic characteristics and NREM/REM sleep appears alternately. Moreover, manual sleep stage scoring depends on not only temporally local features, but also the epochs before and after the current epoch \cite{dong2017mixed}. Recurrent neural networks (RNNs), particularly those using Bidirectional Long Short-Term Memory (BLSTM) hidden units, are powerful models for learning from sequence data \cite{gers2002learning,lipton2015critical}. They are capable of capturing long-range dependencies, which make RNNs quite suitable for modeling sleep data. Inspired by this, we apply a BLSTM-based RNN architecture for classification.

In this paper, we develop an automatic sleep stage classification algorithm using heart rate and wrist actigraphy derived from the wearable device. The proposed method consists of two phases: multi-level feature learning and RNN-based classification. In the feature extraction phase, contrary to traditional methods that extract specific hand-engineered features using much prior domain knowledge, we aim to obtain main information of sleep data. Low-level features (mean value, DCT, cepstral analysis) are extracted from raw signals and mid-level representations are explored. In the classification phase, the BLSTM-based RNN architecture is employed to learn temporally sequential patterns.

The contributions of our algorithm include:

\begin{enumerate}
	\item A complete sleep stage classification solution specially designed for wearable devices is proposed which leverages heart rate and wrist actigraphy.
	\item We introduce mid-level feature learning into sleep staging area, and demonstrate its effectiveness.
	\item The feasibility of using RNN to model sleep signals is verified.
\end{enumerate}

\section{Methods}
\begin{figure*}[!t]
\captionsetup{labelsep=space}
\includegraphics[width=1\linewidth]{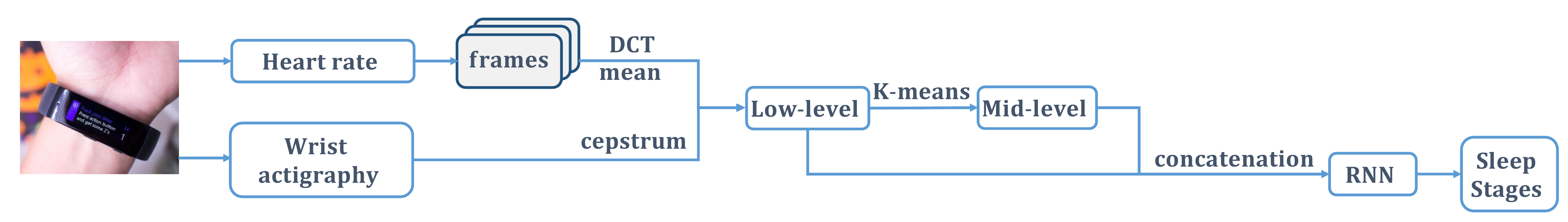}
\caption{The flowchart of the proposed method. The method consists of two phases: multi-level feature learning and RNN-based classification. In the first phase, low-level features are extracted from heart rate and wrist actigraphy signals. Then mid-level features are obtained based on low-level ones. Combining two levels of features, we arrive at the final representations. In the second phase, a BLSTM-based RNN architecture is applied for classification. The obtained features serve as inputs to the network and predictions of sleep stages are finally output by RNN.}
\label{fig_sim}
\end{figure*}

Our method consists of two phases: multi-level feature learning and RNN-based classification. The flowchart of the whole algorithm is shown in Fig. \ref{fig_sim}.

\subsection{Feature extraction}
\subsubsection{Low-level feature extraction}
Low-level features reflect intuitive properties of underlying activities in signals. We first divide each heart rate and actigraphy records into 30s epochs synchronizing in time with PSG classification results. For each epoch, low-level features of heart rate and actigraphy are extracted, respectively. Then they are combined to learn mid-level features.

Previous research efforts in HRV analysis have shown that both time domain and frequency domain measures contain valuable information related to sleep staging \cite{otzenberger1998dynamic}. Therefore, both temporal and frequency properties of heart rate are considered. To make features more representative in context, the feature extraction procedure is carried out based on the sliding window technique by sampling overlapping frames from signal streams. We extract features in a frame which includes 10 epochs centered around the current one. Features are extracted within each epoch and then concatenated to form the final features of the current sleep epoch. First, we derive RR intervals from heart rate,
\begin{equation}
    RR = 60/HR,
\end{equation}
in which RR refers to RR intervals and HR refers to heart rate.

Time domain analysis, which reveals the overall level and variation of the signal by statistical evaluation, is one of the most common approaches to characterize HRV. We compute the mean RR intervals of each epoch in the frame to constitute a mean value vector.

Frequency domain features of RR intervals can capture the activity of ANS. Compared with Discrete Fourier Transform (DFT), Discrete Cosine Transform (DCT) shows better performance with respect to energy concentration \cite{ahmed1974discrete}. We adopt DCT to RR intervals through which dominant frequency components in each epoch are procured to form the frequency feature vector. To measure the fluctuation of RR intervals in frequency domain, we also calculate the first and the second order difference of the dominant frequency components vector acquired above. Then the zero order (the dominant frequency components vector), first order, and second order difference frequency components are joined together as frequency domain features of heart rate.

Humans usually keep still most of the time during sleep, and body movement tends to be transient, which means that a wide range of context information is not needed for actigraphy. Meanwhile, the sampling rate of actigraphy in our work is high enough to capture details. Therefore, actigraphy features are extracted only within the current epoch. The cepstral analysis, which is widely used in action recognition area \cite{kang1995application}, has also been implemented in sleep study to assess body movement \cite{kortelainen2010sleep}. In this study, we calculate the first order difference of the actigraphy along three axes, respectively. Then the dominant cepstrum components of the aforementioned difference in each axis are concatenated to form the actigraphy feature vector.

\subsubsection{Mid-level feature learning}
Mid-level feature learning methods are widely used in various kinds of pattern recognition tasks and give a nice performance boost \cite{barthelemy2013multivariate,liu2016dictionary,wang2013biomedical}. Compared with low-level feature extraction which involves analyzing properties in time and frequency domains, mid-level feature learning pays more attention to analyzing compositions and exploring inherent structure of signals \cite{xu2017learning}. It can be assumed that sleep is comprised of different compositions. Weights of each composition vary in different stages. Thus bag-of-words (BOW), a kind of dictionary learning method is quite appropriate for obtaining mid-level sleep representations. In this work, we implement BOW based on low-level features of both heart rate and actigraphy signals.

The dictionary is first constructed upon low-level features from the training set through $K$-means algorithm \cite{coates2012learning,gunecs2010efficient}. $K$ clusters are thus generated, and each cluster center represents one composition. The dictionary is composed of these cluster centers and forms the whole sleep structure. We define a set of samples as \{$\bm{x}_{1}$, $\bm{x}_{2}$, \ldots, $\bm{x}_{n}$\}, $\bm{x}_{i}\in\mathbf{R}^{d\times1}$, $i\in\{1, 2, \ldots, n\}$ and each sample $\bm{x}_{i}$ is related to an index $z_{i}\in\{1, 2, \ldots, K\}$. If $z_{i} = j\in\{1, 2, \ldots, K\}$, $\bm{x}_{i}$ belongs to the $j$-th cluster. The center of the $j$-th cluster is denoted as
\begin{equation}
    \left.{\bm{m}_{j}=\sum_{i=1}^{n}1\{z_{i}=j\}\bm{x}_{i}}\middle/{\sum_{i=1}^{n}1\{z_{i}=j\}}\right.$, $\bm{m}_{j}\in\mathbf{R}^{d\times1},
\end{equation}
in which $\bm{m}_{j}$ refers to a word (i.e. sleep composition) and $j$ refers to the corresponding index in the dictionary (i.e. the whole sleep structure). After clusters are first built using $K$-means clustering algorithm, the Euclidean distances between low-level features and each cluster are computed as mid-level features which manifest different impacts of each composition.

We concatenate low-level and mid-level features as final features. Final features are then normalized with Z-score strategy.
\subsection{Recurrent neural networks}
\begin{figure*}[ht]
\captionsetup{labelsep=space}
    \includegraphics[width=1\linewidth]{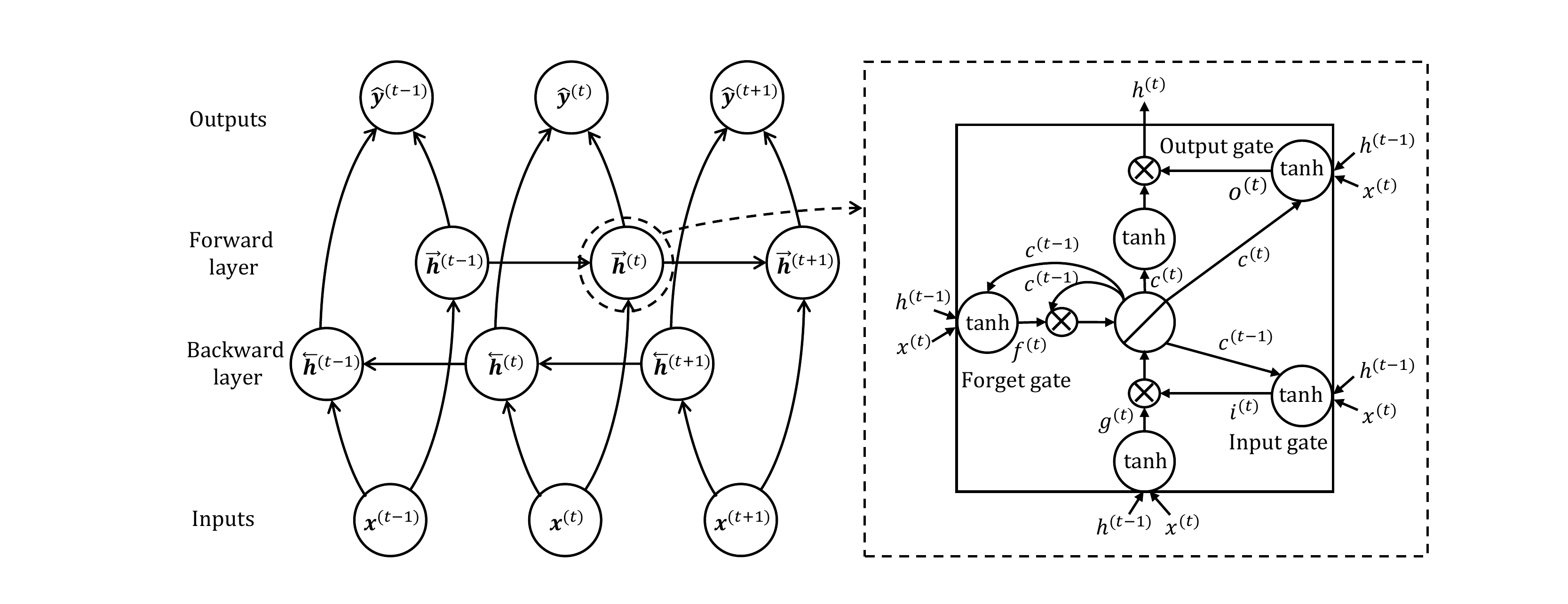}
    \caption{Illustration of the BLSTM architecture. The left side is the overall view. There exist two hidden layers with opposite directions. Each hidden unit is a LSTM memory cell. The right side shows the detail of the LSTM memory cell.}
    \label{lstm_brnn}
\end{figure*}

Recurrent neural networks are suitable models for sequential data and have gained great success in numerous sequence learning domains, including natural language processing, speech, and video \cite{lipton2015critical}. RNNs can capture the dynamics of sequences through loops in the network of units, which means that RNNs can utilize previous information for the present task. In fact, the long-term information utilization is key to sleep staging. For example, there always exists a long-term memory of heart rate in REM sleep and wake stage but weak long-term correlation in NREM sleep \cite{bunde2000correlated}. Given that sleep is an inherently dynamic and long-term dependency process, it seems natural to consider RNNs as a potentially effective model.

Generally, we define input units $\bm{x}=(\bm{x}^{(1)}, \ldots, \bm{x}^{(T)})$, hidden units $\bm{h}=(\bm{h}^{(1)}, \ldots, \bm{h}^{(T)})$ and output units $\bm{\hat{y}}=(\bm{\hat{y}}^{(1)}, \ldots, \bm{\hat{y}}^{(T)})$, where each input $\bm{x}^{(t)}\in\mathbf{R}^{D\times1}$, $\bm{\hat{y}}$ is the hypothesis of the true label $\bm{y}$ and each output $\bm{\hat{y}}^{(t)}\in \left[0,1\right]^{M}$. Here, $D$ refers to the dimension of final features, $t$ refers to the $t$-th sleep epoch, $T$ refers to the total number of sleep epochs and $M$ refers to the total classes of the sleep stage. RNNs have the form of repeating modules which make them learn from the past. More specifically, the input of the hidden unit contains both the current output of the input unit and the previous output of the hidden unit. The hidden unit and the output unit get updated by iterating the following equation from $t=1$ to $T$:
\begin{equation}
    \bm{h}^{(t)} = \mathcal{H}(W_{xs}\bm{x}^{(t)}+W_{sy}\bm{h}^{(t-1)}+\bm{b}_{h}),
\end{equation}
\begin{equation}
    \bm{\hat{y}}^{(t)}= \mathcal{S}(W_{hy}\bm{h}^{(t)}+\bm{b}_{y}),
\end{equation}
in which $W$ refers to weight matrices and $b$ refers to bias vectors (with superscripts denoting time steps and subscripts denoting layer indices). $\mathcal{H}$ denotes the hidden layer function. In traditional RNNs, $\mathcal{H}$ is usually a sigmoid, tanh or rectified linear unit (ReLU) function that is applied element-wise. $\mathcal{S}$ denotes the output layer function. Note that we utilize RNNs for conducting multiclass sequence classification with $M$ alternative classes, we apply a softmax function to output probabilities of the $M$ classes.

However, this original version of RNNs is not able to learn the ``long-term dependencies'' due to the ``gradient varnishing'' problem \cite{bengio1994learning}. The introduction of Long Short-Term Memory units (LSTM) \cite{hochreiter1997long} which replaces the traditional $\mathcal{H}$ function, enables RNNs to memorize information for long periods of time. The version of LSTM memory cells that we utilize is with forget gates \cite{gers1999learning} and peephole connections \cite{gers2002learning}. The right part of Fig. \ref{lstm_brnn} illustrates the structure of the LSTM memory cell. The key point of LSTM is the cell state $c$ that capacitates RNNs to memorize by removing or adding information to it. This manipulation is mainly regulated by three modules, namely the input gate ($\bm{i}$), forget gate ($\bm{f}$), output gate ($\bm{o}$). LSTM proceeds by the following functions:
\begin{equation}
    \bm{i}^{(t)} = \sigma(W_{xi}\bm{x}^{(t)}+W_{hi}\bm{h}^{(t-1)}+W_{ci}\bm{c}^{(t-1)}+\bm{b}_{i}), \qquad\quad\
\end{equation}
\begin{equation}
    \bm{f}^{(t)} = \sigma(W_{xf}\bm{x}^{(t)}+W_{hf}\bm{h}^{(t-1)}+W_{cf}\bm{c}^{(t-1)}+\bm{b}_{f}), \qquad\ \,
\end{equation}
\begin{equation}
    \bm{c}^{(t)} = \bm{f}^{(t)}\odot \bm{c}^{(t-1)}+\bm{i}^{(t)}\odot \phi(W_{xc}\bm{x}^{(t)}+W_{hc}\bm{h}^{(t-1)}+\bm{b}_{c}),
\end{equation}
\begin{equation}
    \bm{o}^{(t)} = \sigma(W_{xo}\bm{x}^{(t)}+W_{ho}\bm{h}^{(t-1)}+W_{co}\bm{c}^{(t)}+\bm{b}_{o}), \qquad\qquad
\end{equation}
\begin{equation}
    \bm{h}^{(t)} = \phi(\bm{c}^{(t)})\odot \bm{o}^{(t)}, \qquad\qquad\qquad\qquad\qquad\qquad\qquad\ \ \,\,
\end{equation}
in which $\sigma$ is an element-wise application of the logistic sigmoid function, $\phi$ is an element-wise application of the tanh function and $\odot$ denotes element-wise multiplication.

\begin{figure*}[ht]
\captionsetup{labelsep=space}
    \centering
    \includegraphics[width=0.8\linewidth]{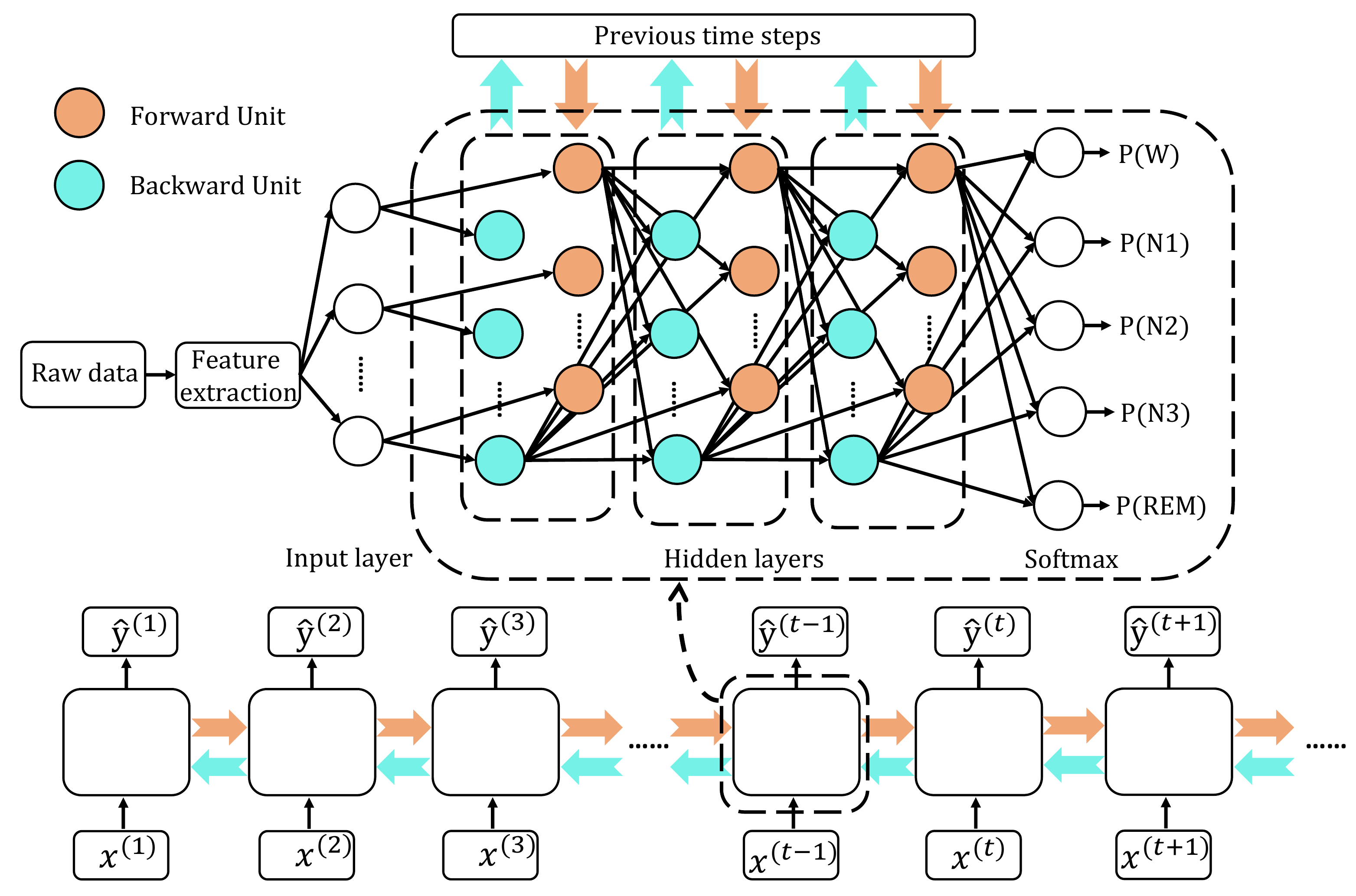}
    \caption{Illustration of the proposed RNN classifier. The upper part is the detailed structure of the network. In order to express concisely, we omit the full connection between layers. The lower part depicts the workflow of the RNN. For a certain sleep epoch, data are processed in two opposite directions. The output layer predicts the sleep stage of the current epoch.}
    \label{overall_network}
\end{figure*}

Apart from the gradient problem, traditional RNNs also have a defect that they are only able to make use of previous information. Since sleep is a kind of sequential signal that shows periodic variations over time, combining context information from both past and future may contribute to better modeling. Bidirectional RNNs (BRNNs) \cite{schuster1997bidirectional} meet this requirement by deploying two separate hidden layers. As shown in Fig. \ref{lstm_brnn}, both hidden layers are connected to input and output, but the processing of data in two hidden layers are of opposite directions: one has recurrent connections from past time steps, the other from the future. The output layer does not get updated until both hidden layers finish computation. The following three equations describe BRNNs:
\begin{equation}
    \overrightarrow{\bm{h}}^{(t)} = \mathcal{H}(W_{x\overrightarrow{h}}\bm{x}^{(t)}+W_{\overrightarrow{h}\overrightarrow{h}}\overrightarrow{\bm{h}}^{(t-1)}+\bm{b}_{\overrightarrow{h}}),
\end{equation}
\begin{equation}
    \overleftarrow{\bm{h}}^{(t)} = \mathcal{H}(W_{x\overleftarrow{h}}\bm{x}^{(t)}+W_{\overleftarrow{h}\overleftarrow{h}}\overleftarrow{\bm{h}}^{(t+1)}+\bm{b}_{\overleftarrow{h}}),
\end{equation}
\begin{equation}
    \bm{\hat{y}}^{(t)} = \mathcal{S}(W_{\overrightarrow{h}y}\overrightarrow{\bm{h}}^{(t)}+W_{\overleftarrow{h}y}\overleftarrow{\bm{h}}^{(t)}+\bm{b}_{y}),
\end{equation}
in which $\overrightarrow{\bm{h}}$ and $\overleftarrow{\bm{h}}$ represent the hidden layers in the forwards and backwards directions, respectively.

As mentioned above, LSTM introduces a new type of unit that constitutes the hidden layer, and BRNNs highlight the strategy of network connections. Combining the above two ideas, bidirectional LSTM (BLSTM) \cite{graves2005framewise,graves2013speech} is formed, which can exploit long-range information in both input directions.

In this paper, we train a BLSTM-based RNN model with multiple hidden units. We also evaluate how RNN can benefit from the use of deep architectures. Specifically, by stacking multiple recurrent hidden layers on top of each other, the way that conventional deep networks do, we arrive at the deep BLSTM. The structure of the deep BLSTM is shown in Fig. \ref{overall_network}. Assuming there are $N$ hidden layers with hidden units of BLSTM, the hidden sequences $\bm{h}^{(t)}_{n}$ are computed from $n=1$ to $N$ and $t=1$ to $T$ as below:
\begin{equation}
    \bm{h}^{(t)}_{n}=\mathcal{H}(W_{h_{n-1}h_{n}}\bm{h}_{n-1}^{(t)}+W_{h_{n}h_{n}}\bm{h}_{n}^{(t-1)}+\bm{b}_{h,n}).
\end{equation}
Thus the output $\bm{\hat{y}}^{(t)}$ can be defined as:
\begin{equation}
    \bm{\hat{y}}^{(t)}=\mathcal{S}(W_{h_{N}y}\bm{h}_{N}^{(t)}+\bm{b}_{y}).
\end{equation}
According to the softmax output, cross entropy function is used as the loss function that we optimize:
\begin{equation}
    loss(\bm{\hat{y}},\bm{y})=-\frac{1}{T}\sum_{t=1}^{t=T}(\bm{y}^{(t)}\cdot log(\bm{\hat{y}}^{(t)})+(1-\bm{y}^{(t)})\cdot log(1-\bm{\hat{y}}^{(t)})),
\end{equation}
in which $\bm{y}^{(t)}$ refers to the true label at time step $t$.
\section{Experiments and Results}
\subsection{Subjects and data}

\subsubsection{Subjects}
Sleep recordings (heart rate and wrist actigraphy) used in this study were obtained from the General Hospital of the Air Force, PLA, Beijing, China. The study recruited 40 healthy subjects (30 males and 10 females) with an age range of 19-64 years old. None of them took drugs or medications that could affect sleep before the experiment. During sleep, subjects were equipped with both the wearable device and the PSG system. The former was used to collect heart rate and wrist actigraphy. The latter was used for labeling sleep data to obtain ``golden standard''. From a practical perspective, whether the patient is in a resting state or has just finished exercise is unknown, and changes in heart rate under two different conditions are not quite the same. Therefore, to cope with the aforementioned issue, we conducted experiments in two groups: the resting group, recordings of which are gathered in resting state, and the comprehensive group, which aims to simulate sleep in the actual situation. 28 recordings were included in the resting group. The remaining 12 subjects were asked to do adequate exercise one to two hours before sleep. All recordings are combined to form the comprehensive group. Eventually, we obtained 39 sleep recordings, with the collection of one recording in non-resting state failed.

\subsubsection{Data collection}
The wearable device that we utilized was Microsoft Band 1. Subjects were required to get plenty of exercise during the daytime and not to take a nap at noon. The band was worn on the right wrist. The placement of the band and PSG electrodes were completed before 9: 00 p.m. Subjects were awakened by the doctor at 6: 00 the next morning. The band was fully charged and synchronized with the PSG system. After the subject awakened, heart rate and triaxial wrist actigraphy were exported and the band was recharged. It is worth noting that the band exported heart rate by converting the RR interval at each detected R peak. The actigraphy was stored in the unit of gravity (g) and sampled at 32 Hz.

\subsubsection{Ground truth}
\begin{figure}[!t]
\captionsetup{labelsep=space}
\centering
\includegraphics[width=8.5cm]{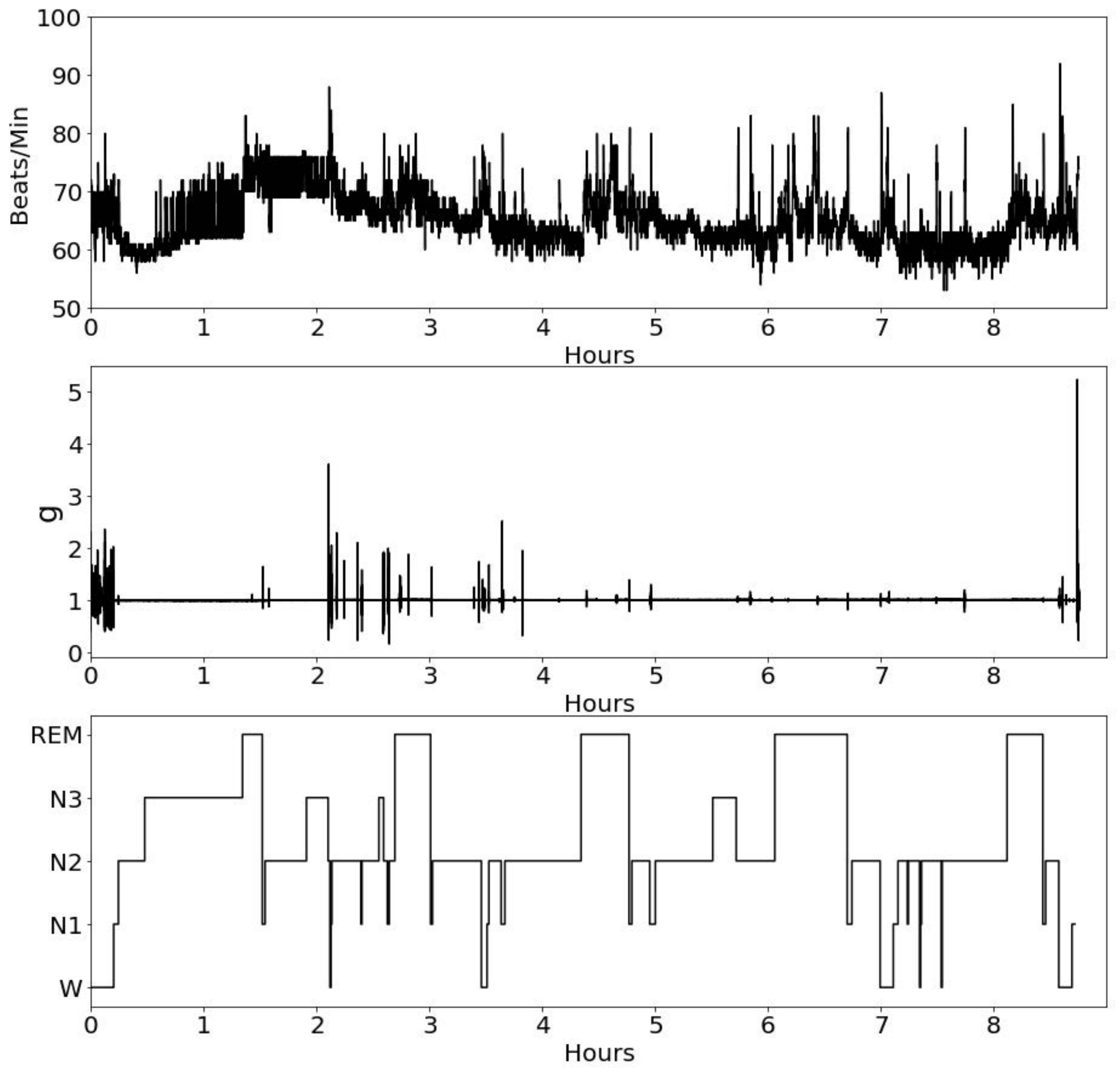}
\caption{Illustration of the recording. The top one represents an overnight heart rate signal. The middle one represents an actigraphy signal (Here, to save space, we integrate the actigraphy in X, Y, and Z axes). The bottom one represents the corresponding sleep stages.}
\label{raw}
\end{figure}

PSG data were acquired using Compumedics E-64. All subjects slept at the same sleep lab and used the same set of PSG equipment. The sleep physician assigned one of the five stages (W, N1, N2, N3, and REM) to the overnight sleep at each epoch according to the standard protocol of the AASM guideline. To ensure accuracy of ground truth, each subject's PSG recording was scored by 5 physicians from the General Hospital of the Air Force independently. Then we adopted the voting strategy to obtain final staging results. For the cases in which multiple stages tied for the most votes, we followed the stage of the previous epoch. An example of recordings is shown in Fig. \ref{raw}. Details of subject demographics are listed in Table \ref{demographics}.

\begin{table}[htbp]
\captionsetup{labelsep=space}
\caption{\\SUBJECT DEMOGRAPHICS}
\label{demographics}
\centering
\begin{threeparttable}
\begin{tabular}{p{1.3cm}|p{1.38cm}p{1.36cm}|p{1.38cm}p{1.36cm}}
\hline
\hline
& \multicolumn{2}{c|}{Resting group} & \multicolumn{2}{c}{Comprehensive group} \\
 \hline
 Parameter & Mean$\pm$Std & Range & Mean$\pm$Std & Range \\
 \hline
 Num &     28      &          &      39     &          \\
 M/F &       20/8    &          &     30/9      &          \\
 Age ($y$) & 26.25$\pm$7.77 & 19-48 & 27.72$\pm$10.12 & 19-64 \\
 BMI & 21.81$\pm$2.66 & 17.07-29.05 & 21.81$\pm$2.68 & 17.07-29.05 \\
 TRT ($h$) & 7.80$\pm$0.52 & 6.92-8.80 & 7.86$\pm$0.54 & 6.82-8.80 \\
 W ($\%$) & 19.14$\pm$8.80 & 4.81-40.24 & 19.37$\pm$9.13 & 4.81-40.24 \\
 N1 ($\%$) & 5.79$\pm$3.20 & 1.73-14.18 & 6.05$\pm$3.43 & 1.73-15.09 \\
 N2 ($\%$) & 45.00$\pm$9.48 & 26.04-62.31 & 46.14$\pm$8.84 & 26.04-62.31 \\
 N3 ($\%$) & 14.86$\pm$9.57 & 0.00-42.23 & 13.47$\pm$8.73 & 0.00-42.23 \\
 REM ($\%$) & 15.20$\pm$4.47 & 6.15-24.30 & 14.96$\pm$4.15 & 6.15-24.30 \\
 \hline
 \hline
\end{tabular}
\begin{tablenotes}\footnotesize
\item[] {Num = number of recordings, M/F = male/female, BMI = body mass index, TRT = total recording time.}
\end{tablenotes}
\end{threeparttable}
\end{table}

\subsection{Performance Evaluation}
In this study, a subject independent cross validation (CV) procedure is considered in order to present unbiased performance of the RNN model. More precisely, an eight-fold CV is conducted. On each iteration, we use six portions for training, one portion for validation, and one portion for testing. Finally, testing results of each iteration are averaged to form the overall performance of RNN classifiers. We execute the CV for three rounds and calculate average results.

Considering that the distribution of the five sleep stages is severely unbalanced, to adapt this characteristic, weighted precision (P), recall (R) and $F_{1}$ score ($F_{1}$) are selected to evaluate the performance of our method. Evaluation measures are defined as:
\begin{equation}
    P = \sum_{i}{\omega_{i}\cdot}TP_{i}/\left({TP_{i}+FP_{i}}\right),
\end{equation}
\begin{equation}
    R = \sum_{i}{\omega_{i}\cdot}TP_{i}/\left({TP_{i}+FN_{i}}\right),
\end{equation}
\begin{equation}
    F_{1} = \sum_{i}{2\cdot\omega_{i}\cdot}P_{i}\cdot{R_{i}}/\left({P_{i}+R_{i}}\right),
\end{equation}
in which $i$ refers to the stage category and $\omega_{i}$ is the proportion of the $i$-th stage class in all classes. $TP$ is the number of true positives, $FP$ is the number of false positives, $TN$ is the number of true negatives and $FN$ is the number of false negatives (Here, we omit the subscript $i$.).

\subsection{Experiments}
In order to fully explore the property of the proposed approach, we conduct the following experimental procedures: (1) we exhibit our algorithm's results, and compare them with the existing method; (2) we prove the effectiveness of mid-level feature learning; (3) we provide a comparison between a BLSTM-based RNN and two frequently-used classifiers for classification; (4) we explore the sensitivity of parameters in the feature extraction process; (5) we evaluate the performance of RNN models with different hidden layer width, depth, and unit types. As W and N1 stages are morphologically similar and hard to separate, we also conduct experiments that combine W and N1 into one stage, resulting in 4 classes of classification. Meanwhile, the resting group and comprehensive group are both experimented on, respectively.

\subsubsection{Comparison with the existing method}
We first describe parameter settings of our procedure and present the results. Then we compare our algorithm with a similar one described in \cite{xiao2013sleep} which classifies sleep stages based on HRV and RF.

In low-level feature extraction, for frequency domain features of heart rate, the first 5 frequency components of each epoch are collected to make up frequency features. Consequently, we obtain a 120-dimension feature (40 dimensions for each axis, respectively). For actigraphy features, we join the first 30 cepstrum components in each axis to shape into a 90-dimension actigraphy feature. Based on the above low-level features, we extract mid-level features. The size of the dictionary, $K$, is set to 300. The final features are formed by the concatenation of low-level and mid-level features, holding a dimension of 520. The RNN architecture that we design for classification is a three-layer structure with the number of input layer units equal to the dimension of the final feature vector, one hidden layer of 400 BLSTM cells and the output layer of 5 units. RNN is trained by using stochastic gradient descent \cite{bousquet2008tradeoffs} with the learning rate of $10^{-6}$ and weights are randomly initialized with Gaussian distribution of $(0, 0.1)$. To be more generalized, the network is trained with the Gaussian weighted noise $(\sigma = 0.005)$. We implement the proposed RNN architecture under the CURRENNT framework \cite{weninger2015introducing}.

Noting that the existing study used different datasets for evaluation, it is not appropriate to make a comparison by directly using the results in their paper. Here we implement the baseline method using RR intervals in our dataset. The baseline method extracts 41-dimension hand-engineered features based on HRV in time domain (8 dimensions), frequency domain (20 dimensions) and nonlinear analysis (13 dimensions) and these features are then trained and tested through RF. To make a fair comparison, we also implement our proposed algorithm without actigraphy. Moreover, we add our extracted actigraphy features to 41-dimension HRV features to perform the baseline method. Both the resting group and the comprehensive group are used to evaluate experiments for 5-class classification. Table \ref{baseline} shows the results.

\begin{table}[htbp]
\captionsetup{labelsep=space}
\caption{\\COMPARISON WITH THE EXISTING METHOD FOR FIVE-CLASS CLASSIFICATION}
\label{baseline}
\begin{threeparttable}
\begin{tabular}{p{2.26cm}|p{0.54cm}p{0.54cm}p{0.54cm}|p{0.54cm}p{0.54cm}p{0.54cm}}
\hline
\hline
 \multirow{2}{*}{Method}& \multicolumn{3}{c|}{RG} & \multicolumn{3}{c}{CG} \\
\cline{2-7}
  & P & R & $\rm{F_{1}}$ & P & R & $\rm{F_{1}}$ \\
\hline
Baseline (HR) & 50.2 & 47.5 & 43.2 & 48.8 & 47,2 & 42.1 \\
Baseline (HR+Act) & 53.5 & 51.7 & 46.0 & 52.2 & 51.9 & 45.5 \\
Proposed (HR) & 53.9 & 56.0 & 53.2 & 53.2 & 55.5 & 52.8 \\
Proposed (HR+Act) & \textbf{58.0} & \textbf{60.3} & \textbf{58.2} & \textbf{58.5} & \textbf{61.1} & \textbf{58.5} \\
\hline
\hline
\end{tabular}
\begin{tablenotes}\footnotesize
\item[] {RG = resting group, CG = comprehensive group, HR = heart rate, Act = actigraphy.}
\end{tablenotes}
\end{threeparttable}
\end{table}
It can be observed from Table \ref{baseline} that our method surpasses the baseline both with and without actigraphy features. The best performances, weighted precision, recall and $F_{1}$ score of 58.0\%, 60.3\%, and 58.2\% in the resting group and 58.5\%, 61.1\%, and 58.5\% in the comprehensive group are achieved through our method using heart rate combined with actigraphy. Compared with the hand-engineered feature extraction method in the baseline which is overly dependent on expert knowledge, our feature learning method aims to obtain main information of signals and RNN is able to further refine features. As little prior domain knowledge is used in our method, the method has the potential to generalize sleep disorder detection. Approximately, the differences of results between the resting group and comprehensive group using our method are smaller than that using the baseline. Since sleep data in the comprehensive group are more diverse, it shows the robustness of RNN.

Furthermore, it can be noticed that the performances of both methods are improved when actigraphy features are considered, which suggests that body movement during sleep contains useful information for sleep stage classification.

\subsubsection{Effectiveness of mid-level learning}
To demonstrate the effectiveness of mid-level feature learning, we design a contrast experiment in which low-level features directly serve as the input to RNN without mid-level feature learning. Parameter settings remain the same. We conduct experiments with the resting group for 5-class classification. 1 to 4 hidden layers of RNN are implemented.

As shown in Table \ref{mid-level}, the performance is improved significantly when mid-level features are involved. By building a dictionary, we obtain the spacial distribution of sleep compositions and explore inherent structures, which can describe sleep in a more representative way.

\begin{table}[htbp]
\captionsetup{labelsep=space}
\caption{\\COMPARISON OF RESULTS WITH AND WITHOUT MID-LEVEL LEARNING WITH THE RESTING GROUP FOR 5-CLASS CLASSIFICATION}
\label{mid-level}
\begin{threeparttable}
\begin{tabular}{p{1cm}|p{0.75cm}p{0.75cm}p{0.75cm}|p{0.75cm}p{0.75cm}p{0.75cm}}
\hline
\hline
\multirow{2}{*}{HL} & \multicolumn{3}{c|}{With mid-level learning} & \multicolumn{3}{c}{Without mid-level learning} \\
\cline{2-7}
  & P & R & $\rm{F_{1}}$ & P & R & $\rm{F_{1}}$ \\
\hline
1 & \textbf{58.0} & \textbf{60.3} & \textbf{58.2} & 52.8 & 54.7 & 52.7 \\
2 & 57.7 & 59.6 & 57.1 & 53.5 & 55.5 & 53.3 \\
3 & 56.3 & 59.6 & 57.1 & 53.1 & 55.7 & 53.1 \\
4 & 55.4 & 58.7 & 56.4 & 52.2 & 54.7 & 51.5 \\
\hline
\hline
\end{tabular}
\begin{tablenotes}\footnotesize
\item[] {HL = number of hidden layers.}
\end{tablenotes}
\end{threeparttable}
\end{table}

\subsubsection{Comparison with various classifiers}
We make a comparison between RNN and two classic classifiers, including Support Vector Machine (SVM) \cite{suykens1999least} and Random Forest (RF). All three classifiers use the same features extracted in the first experiment. The comparison is performed with the comprehensive group for 5-class classification.

The SVM uses the radial basis function (RBF) kernel with the kernel coefficient $gamma$ of 0.0021. Penalty parameter $C$ is set to 1 to regularize the estimation. The shrinking heuristic is also utilized. 600 estimators are set in the RF, and the function to measure the quality of a split is the Gini impurity. The number of features to consider when looking for the best split is denoted as $p$, which is equal to the square root of the number of total features. The minimum number of samples required to split an internal node and be at a leaf node are $p^{1/2}$ and 1.

The results shown in Table \ref{classifiers} indicate that RNN is superior to RF in all three metrics and to SVM in weighted recall and $F_{1}$ score. Despite both SVM and RF being able to deal with high dimension situations, RNN works better at learning long-term dependencies.

\begin{table}[htbp]
\captionsetup{labelsep=space}
\caption{\\COMPARISON OF VARIOUS CLASSIFIERS WITH THE COMPREHENSIVE GROUP FOR 5-CLASS CLASSIFICATION}
\label{classifiers}
\begin{tabular}{p{1.8cm}p{1.8cm}p{1.8cm}p{1.8cm}}
\hline
\hline
Classifier & P & R & $\rm{F_{1}}$ \\
\hline
SVM & \textbf{60.3} & 60.6 & 55.6 \\
RF & 56.5 & 59.2 & 53.3 \\
RNN & 58.5 & \textbf{61.1} & \textbf{58.5} \\
\hline
\hline
\end{tabular}
\end{table}

\subsubsection{Sensitivity of parameters}
This section elucidates the assessment of the variable sensitivity in feature extraction, including the dominant frequency component size of DCT and the dictionary size in mid-level feature learning. We evaluate the sensitivity with the comprehensive group for 4-class classification. Hidden layers from 1 to 4 are employed.

\begin{figure}[!htp]
\label{DCT_Kmeans}
\captionsetup{labelsep=space}
    \centering
    \subfloat[Frequency component size]{\includegraphics[width=0.48\linewidth]{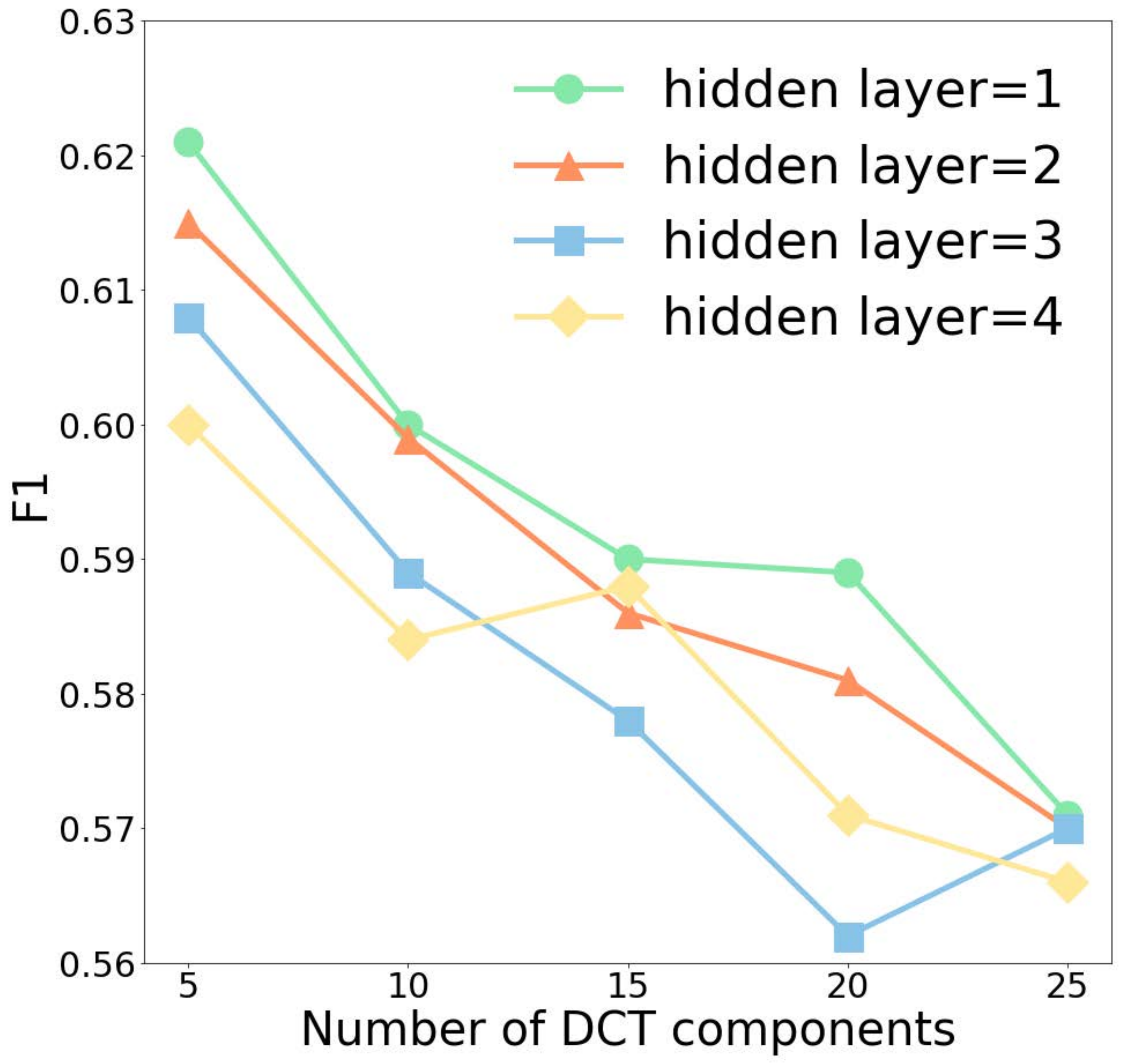}}
    \hspace{0.002\linewidth}
    \subfloat[Dictionary size]{\includegraphics[width=0.48\linewidth]{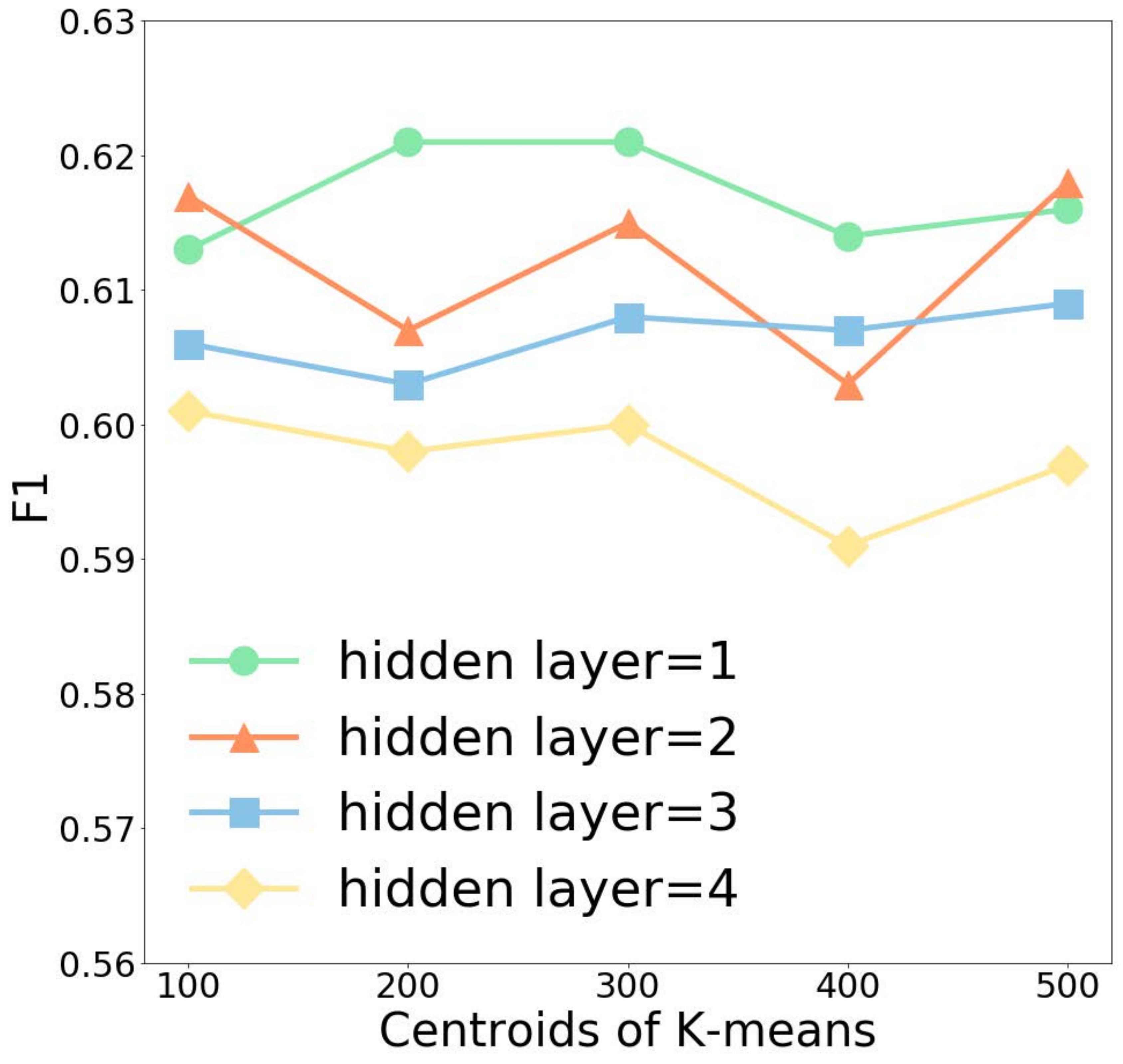}}
    \hspace{0.002\linewidth}
    \caption{The sensitivity of DCT and dictionary size parameters.}
    \label{fig:dct_kmeans}
\end{figure}

The dominant frequency component size ranges from 5 to 25. The results are shown in Table \ref{components39_4} and Fig. \ref{fig:dct_kmeans}(a). It can be seen that performance decreases as frequency component size increases. This may be because the high frequency components are mostly noise and thus impact classification. Even so, the worst result, weighted $F_{1}$ score of 56.2\%, is still higher than that of the baseline method, which demonstrates the effectiveness of the proposed method.

\begin{table*}[htbp]
\centering
\begin{threeparttable}
\captionsetup{labelsep=space}
\captionsetup{justification=centering}
\caption{\\THE EFFECT OF THE DOMINANT FREQUENCY COMPONENT SIZE OF DCT WITH THE COMPREHENSIVE GROUP FOR 4-CLASS CLASSIFICATION}
\label{components39_4}
\begin{tabular}{c ccc ccc ccc ccc}
\hline
\hline
\multirow{3}{*}{Components} & \multicolumn{12}{c}{HL} \\
\cline{2-13}
 & \multicolumn{3}{c}{1} & \multicolumn{3}{c}{2} & \multicolumn{3}{c}{3} & \multicolumn{3}{c}{4} \\
\cline{2-13}
 & P & R & $\rm{F_{1}}$ & P & R & $\rm{F_{1}}$ & P & R & $\rm{F_{1}}$ & P & R & $\rm{F_{1}}$ \\
\hline
5 & \textbf{63.0} & \textbf{63.1} & \textbf{62.1} & 62.1 & 61.9 & 61.5 & 61.7 & 61.9 & 60.8 & 60.6 & 60.8 & 60.0 \\
10 & 60.3 & 61.2 & 60.0 & 60.6 & 60.9 & 59.9 & 60.0 & 60.3 & 58.9 & 59.5 & 60.1 & 58.4 \\
15 & 60.1 & 61.2 & 59.0 & 59.6 & 60.3 & 58.6 & 58.7 & 59.4 & 57.8 & 59.9 & 60.3 & 58.8 \\
20 & 59.4 & 60.6 & 58.9 & 58.5 & 59.2 & 58.1 & 57.9 & 58.5 & 56.2 & 57.6 & 59.0 & 57.1 \\
25 & 58.0 & 59.5 & 57.1 & 57.4 & 58.9 & 57.0 & 58.0 & 59.0 & 57.0 & 57.3 & 58.9 & 56.6 \\
\hline
\hline
\end{tabular}
\begin{tablenotes}\footnotesize
\item[] {HL = number of hidden layers.}
\end{tablenotes}
\end{threeparttable}
\end{table*}

We change the dictionary size in mid-level feature learning from 100 to 500. The results are shown in Table \ref{Kmeans39_4} and Fig. \ref{fig:dct_kmeans}(b). The variation trend of performance is small. It can be attributed to as lack of data. Although the increase in dictionary size creates a more detailed description of sleep compositions, the larger size means that there are more parameters to optimize in RNN. Under such conditions, sleep data of around 37,000 epochs are not sufficient for training the RNN model.

\begin{table*}[htbp]
\centering
\captionsetup{labelsep=space}
\captionsetup{justification=centering}
\begin{threeparttable}
\centering
\caption{\\THE EFFECT OF DICTIONARY SIZE ON THE MID-LEVEL FEATURE LEARNING WITH THE COMPREHENSIVE GROUP FOR 4-CLASS CLASSIFICATION}
\label{Kmeans39_4}
\begin{tabular}{c ccc ccc ccc ccc}
\hline
\hline
\multirow{3}{*}{K} & \multicolumn{12}{c}{HL} \\
\cline{2-13}
 & \multicolumn{3}{c}{1} & \multicolumn{3}{c}{2} & \multicolumn{3}{c}{3} & \multicolumn{3}{c}{4} \\
\cline{2-13}
 & P & R & $\rm{F_{1}}$ & P & R & $\rm{F_{1}}$ & P & R & $\rm{F_{1}}$ & P & R & $\rm{F_{1}}$ \\
\hline
100 & 61.6 & 62.2 & 61.3 & 62.4 & 62.7 & 61.7 & 60.9 & 61.3 & 60.6 & 60.9 & 61.1 & 60.1 \\
200 & 62.4 & 62.7 & 62.1 & 61.2 & 61.4 & 60.7 & 60.5 & 61.0 & 60.3 & 61.5 & 61.4 & 59.8 \\
300 & \textbf{63.0} & \textbf{63.1} & \textbf{62.1} & 62.1 & 61.9 & 61.5 & 61.7 & 61.9 & 60.8 & 60.6 & 60.8 & 60.0 \\
400 & 62.6 & 62.9 & 61.4 & 60.7 & 61.2 & 60.3 & 61.1 & 61.3 & 60.7 & 60.1 & 60.3 & 59.1 \\
500 & 62.0 & 62.7 & 61.6 & 62.1 & 62.5 & 61.8 & 62.2 & 62.0 & 60.9 & 60.6 & 60.9 & 59.7 \\
\hline
\hline
\end{tabular}
\begin{tablenotes}\footnotesize
\item[] {HL = number of hidden layers.}
\end{tablenotes}
\end{threeparttable}
\end{table*}

\subsubsection{Neural networks configurations}
We carry out extensive experiments to assess the performance of neural networks with different hidden layer width, depth, and unit types. We consider 6 scales of hidden units numbers: 100 to 600, 4 scales of hidden layers: 1 to 4, and 3 types of the neural networks: multilayer perceptron (MLP), LSTM, and BLSTM.

\begin{figure}[ht]
\captionsetup{labelsep=space}
    \centering
    \subfloat[4 classes of Resting group]{\includegraphics[width=0.48\linewidth]{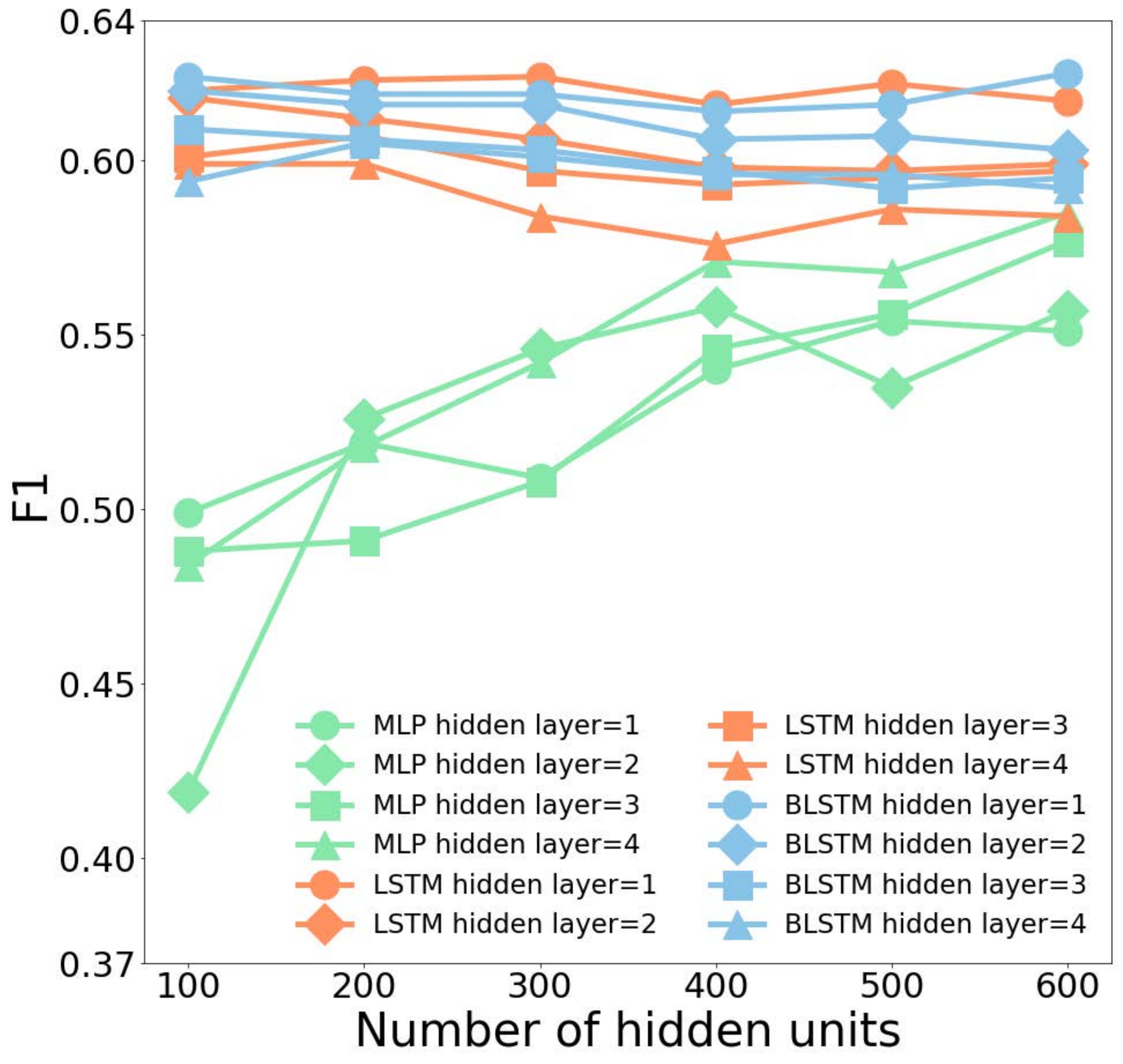}}
    \hspace{0.002\linewidth}
    \subfloat[4 classes of Comprehensive group]{\includegraphics[width=0.48\linewidth]{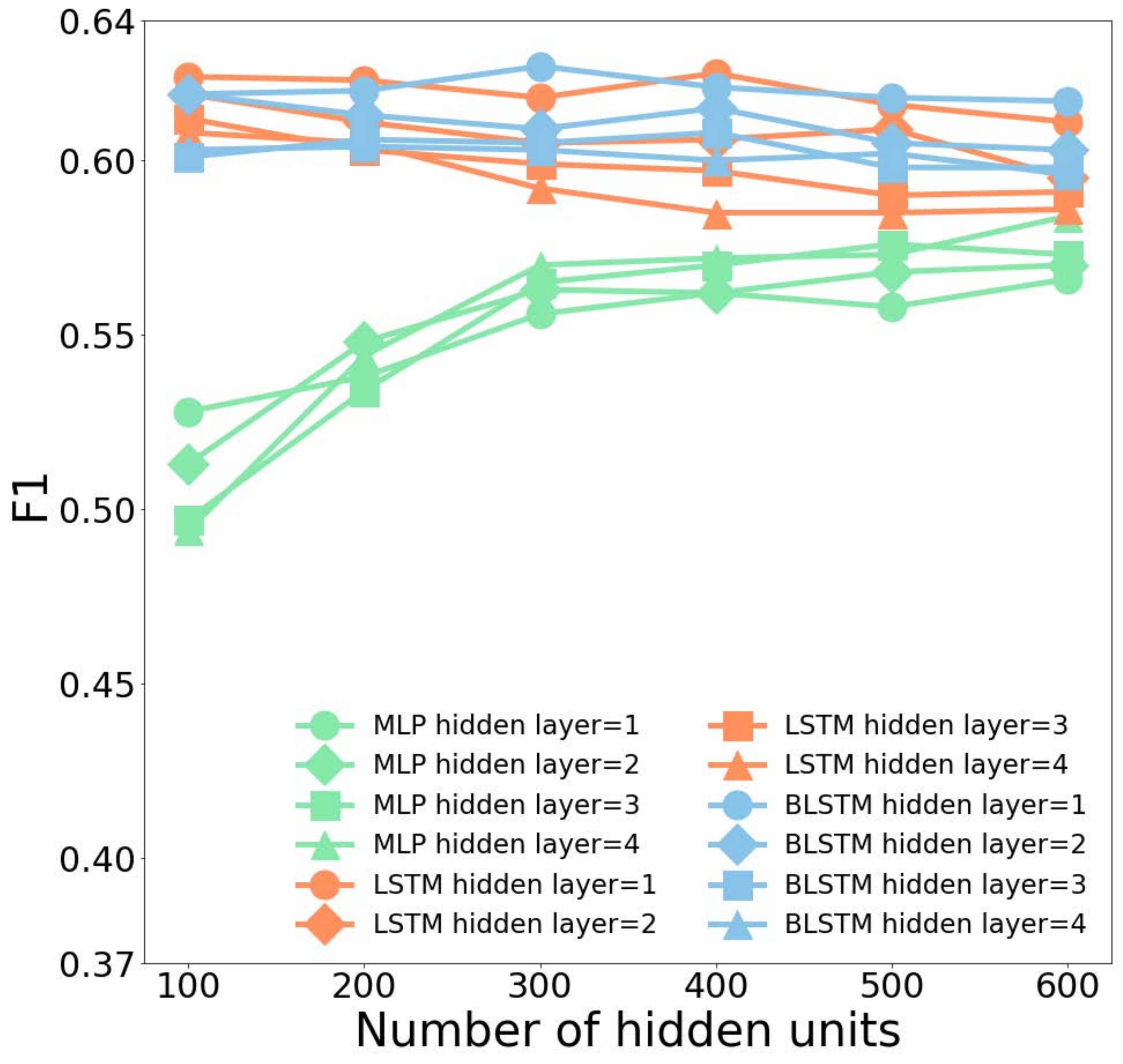}}
    \hspace{0.002\linewidth}

    \subfloat[5 classes of Resting group]{\includegraphics[width=0.48\linewidth]{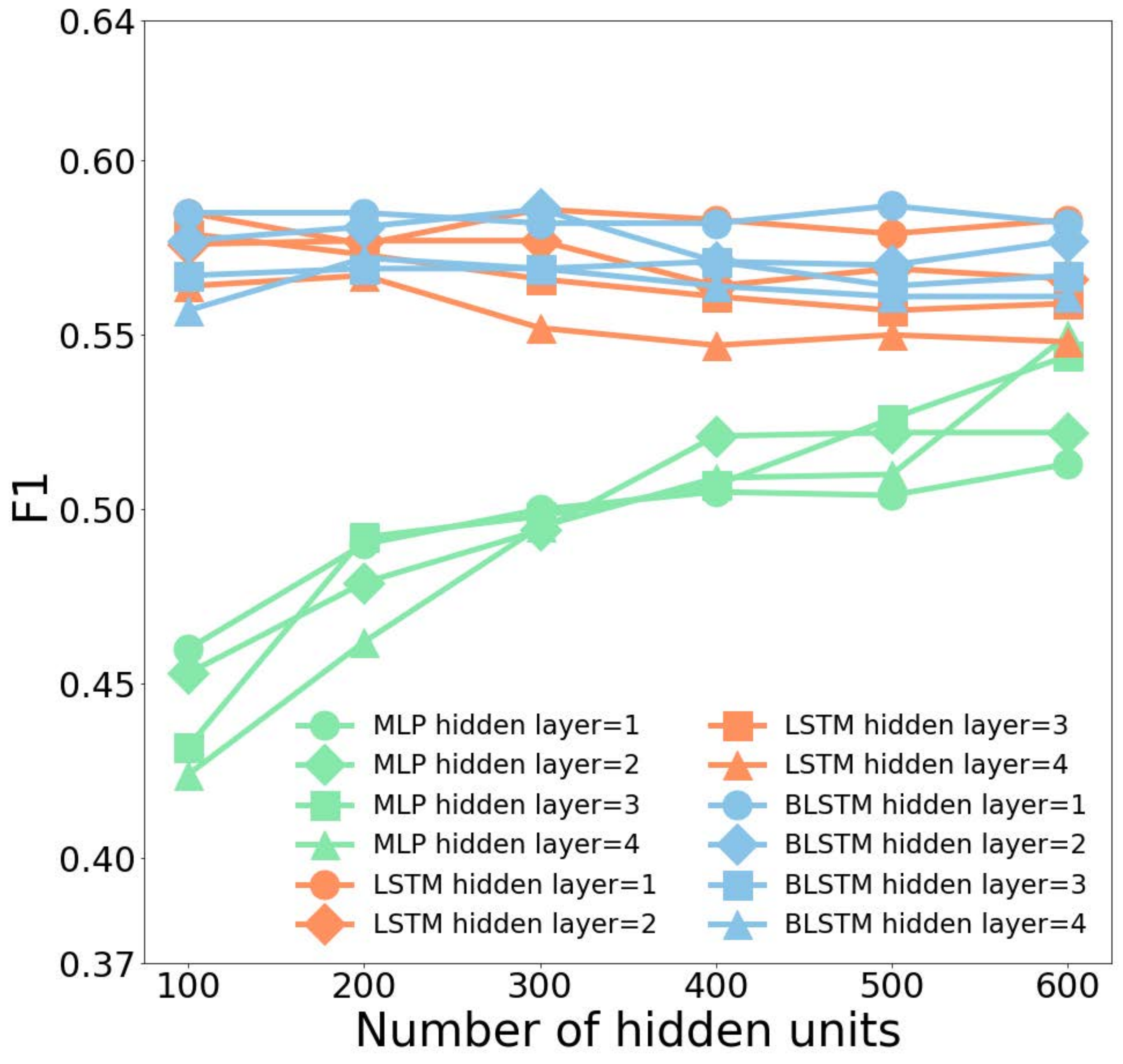}}
    \hspace{0.002\linewidth}
    \subfloat[5 classes of Comprehensive group]{\includegraphics[width=0.48\linewidth]{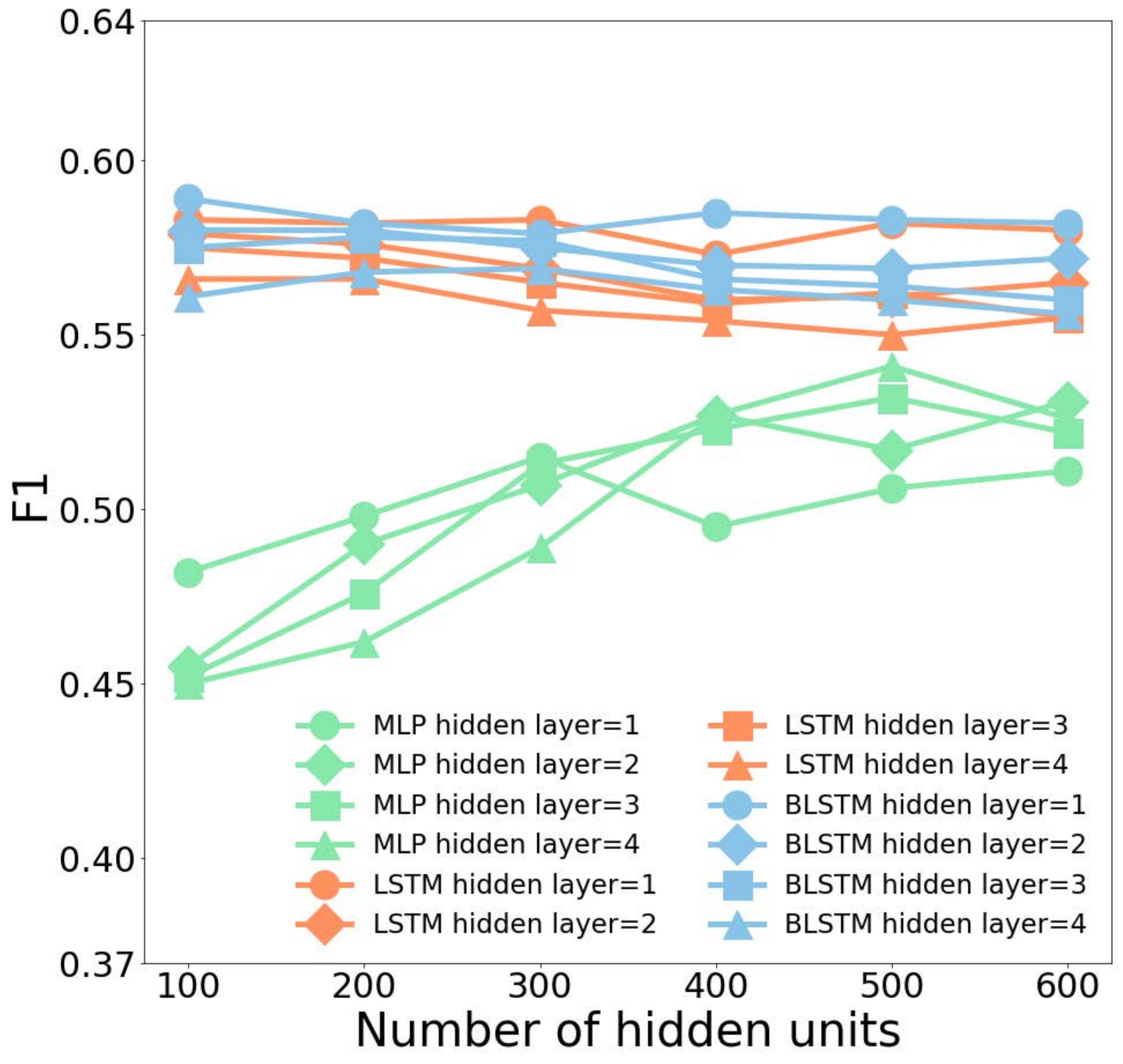}}
    \hspace{0.002\linewidth}
    \caption{Performances of RNN networks with different hidden layers and units.}
    \label{fig:28_39_5}
\end{figure}

Tables \ref{network_4} and \ref{network_5} summarize the results of multiple neural network architectures with the resting group as well as the comprehensive group for 4- and 5-class classification. Fig. \ref{fig:28_39_5} illustrates the general trend of the weighted $F_{1}$ score along with the number of hidden units of four experiment groups. It is obvious that RNNs with LSTM and BLSTM units achieve much better results than MLP in all experiments, which demonstrates the efficacy of RNN to learn the temporal relationships of sleep sequences. BLSTM outperforms LSTM slightly. As the number of hidden units increases, the improvement of performance in MLP is most significant. It is reasonable because MLP models are relatively simple and the collected sleep data are sufficient for optimizing the MLP model with a deeper and wider structure. The performance results for LSTM and BLSTM show a slight decrease as the number of hidden units and layers increase. This may be caused by the mismatch between excessive parameters and relatively sparse data.

\begin{table*}[htbp]
\captionsetup{labelsep=space}
\centering
\begin{threeparttable}
\caption{\\COMPARISON OF DIFFERENT NETWORK ARCHITECTURES FOR 5-CLASS CLASSIFICATION}
\label{network_5}
\begin{tabular}{p{0.7cm} p{0.2cm} p{0.1cm} ccc ccc ccc ccc ccc ccc}
\hline
\hline
      \multirow{3}{*}{Sub.} &  \multirow{3}{*}{HT}& \multirow{3}{*}{HL} & \multicolumn{18}{c}{HU}                                                                                                                   \\
      \cline{4-21}
        &                &  & \multicolumn{3}{c}{100} & \multicolumn{3}{c}{200} & \multicolumn{3}{c}{300} & \multicolumn{3}{c}{400} & \multicolumn{3}{c}{500} & \multicolumn{3}{c}{600} \\
        \cline{4-21}
& & & P & R & $\rm{F_{1}}$ & P & R & $\rm{F_{1}}$ & P & R & $\rm{F_{1}}$ & P & R & $\rm{F_{1}}$ & P & R & $\rm{F_{1}}$ & P & R & $\rm{F_{1}}$ \\
\hline
\multirow{8}{*}{MLP} & \multirow{4}{*}{RG} & 1 &48.7&54.3&46.0&51.5&55.8&49.0&52.0&56.3&50.0&52.2&56.5&50.5&53.0&56.5&50.4&53.5&56.8&51.3\\
                  &                   & 2 &45.6&54.9&45.3&50.8&55.9&47.9&52.8&56.4&49.4&54.8&57.6&52.1&55.4&57.7&52.2&56.3&57.9&52.2\\
                  &                   & 3 &43.4&54.4&43.2&53.3&56.7&49.2&53.1&56.6&49.8&54.8&57.3&50.7&55.7&58.2&52.6&56.2&58.9&54.4\\
                  &                   & 4 &41.4&54.1&42.4&47.2&55.7&46.2&51.4&56.9&49.5&52.2&57.5&50.9&54.7&57.4&51.0&58.2&59.2&55.0\\
                  \cline{2-21}
                  & \multirow{4}{*}{CG} & 1 &50.6&56.3&48.2&52.7&56.8&49.8&53.8&57.7&51.5&53.1&56.8&49.5&53.2&57.2&50.6&54.5&57.5&51.1\\
                  &                   & 2 &47.3&55.5&45.5&52.0&57.0&49.0&54.2&57.4&50.7&55.3&58.4&52.7&55.1&57.9&51.7&56.4&59.0&53.1\\
                  &                   & 3 &44.8&55.5&45.2&50.5&56.2&47.6&56.2&58.1&51.3&55.8&58.4&52.3&56.7&58.7&53.2&55.5&58.9&52.2\\
                  &                   & 4 &44.8&55.7&45.0&49.2&55.8&46.2&49.1&56.9&48.9&56.2&58.3&52.7&56.1&58.9&54.1&55.7&58.6&52.6\\
                  \hline
\multirow{8}{*}{LSTM} & \multirow{4}{*}{RG} & 1 &58.7&61.0&58.5&58.1&60.2&57.6&58.2&60.8&58.6&58.4&60.5&58.3&57.7&60.2&57.9&58.0&60.2&58.3\\
                  &                   & 2 &56.4&60.1&57.6&56.6&60.3&57.7&57.3&60.4&57.7&55.8&58.6&56.4&56.1&59.1&56.9&56.0&59.0&56.6\\
                  &                   & 3 &56.9&60.6&57.9&56.4&59.8&57.3&55.7&59.2&56.6&55.8&58.8&56.1&55.5&58.4&55.7&55.0&58.2&55.9\\
                  &                   & 4 &55.7&59.3&56.4&55.9&59.4&56.7&54.4&57.5&55.2&54.2&57.3&54.7&54.8&57.5&55.0&54.1&57.2&54.8\\
                  \cline{2-21}
                  & \multirow{4}{*}{CG} & 1 & 58.9&61.3&58.3&59.0&61.1&58.2&58.4&61.0&58.3&57.3&59.8&57.3&58.0&60.6&58.2&57.8&60.3&58.0\\
                  &                   & 2 &57.0&60.8&57.9&56.7&60.6&57.6&56.0&59.9&56.9&55.8&58.9&56.0&55.4&58.8&56.1&56.2&59.3&56.5\\
                  &                   & 3 &56.4&60.4&57.5&56.1&60.1&57.2&56.1&59.4&56.5&55.3&59.1&55.9&55.8&59.2&56.2&55.0&58.4&55.5\\
                  &                   & 4 & 55.3&59.3&56.6&55.7&59.7&56.6&54.7&58.5&55.7&54.9&58.6&55.4&54.9&58.1&55.0&55.3&58.5&55.5\\
                  \hline
\multirow{8}{*}{BLSTM} & \multirow{4}{*}{RG} & 1 &58.6&61.0&58.5&58.6&60.8&58.5&58.3&60.6&58.2&58.0&60.3&58.2&58.7&61.0&58.7&58.0&60.3&58.2\\
                  &                   & 2 &57.5&60.2&57.7&57.2&60.5&58.1&58.5&61.0&58.6&57.7&59.6&57.1&56.6&59.3&57.0&57.3&59.8&57.7  \\
                  &                   & 3 &55.6&59.3&56.7&55.7&59.2&56.9&56.0&59.5&56.9&56.3&59.6&57.1&55.7&58.6&56.4&56.3&59.1&56.7 \\
                  &                   & 4 &54.6&58.3&55.7&56.1&59.7&57.2&57.6&59.2&56.9&55.4&58.7&56.4&55.7&58.8&56.1&55.7&58.4&56.1\\
                  \cline{2-21}
                  & \multirow{4}{*}{CG} & 1 &59.0&61.5&58.9&58.4&60.9&58.2&57.6&60.5&57.9&58.5&61.1&58.5&58.2&60.7&58.3&58.1&60.7&58.2\\
                  &                   & 2 &56.8&60.7&58.0&57.1&60.9&58.0&56.3&60.2&57.5&56.7&59.9&57.0&56.9&59.9&56.9&56.7&59.7&57.2\\
                  &                   & 3 &58.6&60.4&57.5&57.0&60.6&57.8&56.8&60.7&57.7&55.9&59.4&56.6&56.4&59.5&56.4&55.5&58.9&56.0\\
                  &                   & 4 &55.1&59.2&56.1&55.5&59.4&56.8&58.0&59.8&56.9&55.0&58.7&56.3&55.3&58.9&56.0&54.7&58.1&55.6\\
                  \hline
                  \hline
\end{tabular}
\begin{tablenotes}\footnotesize
\item[] {Sub. = subjects, RG = resting group, CG = comprehensive group, HT = hidden unit type, HL = number of hidden layers, HU = number of hidden units.}
\end{tablenotes}
\end{threeparttable}
\end{table*}

\begin{table*}[htbp]
\captionsetup{labelsep=space}
\centering
\begin{threeparttable}
\caption{\\COMPARISON OF DIFFERENT NETWORK ARCHITECTURES FOR 4-CLASS CLASSIFICATION}
\label{network_4}
\begin{tabular}{p{0.7cm} p{0.2cm} p{0.1cm} ccc ccc ccc ccc ccc ccc}
\hline
\hline
      \multirow{3}{*}{Sub.} &  \multirow{3}{*}{HT}& \multirow{3}{*}{HL} & \multicolumn{18}{c}{HU}                                                                                                                   \\
      \cline{4-21}
        &                &  & \multicolumn{3}{c}{100} & \multicolumn{3}{c}{200} & \multicolumn{3}{c}{300} & \multicolumn{3}{c}{400} & \multicolumn{3}{c}{500} & \multicolumn{3}{c}{600} \\
        \cline{4-21}
& & & P & R & $\rm{F_{1}}$ & P & R & $\rm{F_{1}}$ & P & R & $\rm{F_{1}}$ & P & R & $\rm{F_{1}}$ & P & R & $\rm{F_{1}}$ & P & R & $\rm{F_{1}}$ \\
\hline
\multirow{8}{*}{MLP} & \multirow{4}{*}{RG} & 1 &53.1&55.8&49.9&55.5&57.1&51.9&53.7&56.8&50.9&57.7&58.5&54.0&58.0&59.3&55.4&58.1&58.8&55.1\\
                  &                   & 2 &44.9&51.0&41.9&56.8&58.2&52.6&57.3&59.4&54.6&59.4&59.8&55.8&57.3&58.9&53.5&58.1&59.6&55.7\\
                  &                   & 3 &52.9&56.6&48.8&51.9&56.6&49.1&56.3&57.4&50.8&57.2&59.4&54.6&59.4&59.8&55.6&61.2&61.0&57.7\\
                  &                   & 4 &49.9&57.1&48.4&56.7&58.2&51.8&56.9&59.2&54.2&60.7&60.6&57.1&59.9&60.5&56.8&60.9&61.2&58.5\\
                  \cline{2-21}
                  & \multirow{4}{*}{CG} & 1 &56.5&58.2&52.8&56.5&58.7&53.8&58.6&59.8&55.6&59.5&60.4&56.2&59.1&60.1&55.8&59.3&60.4&56.6\\
                  &                   & 2 &54.0&57.9&51.3&57.9&59.6&54.8&59.2&60.4&56.3&59.5&60.5&56.2&59.9&60.7&56.8&60.1&60.9&57.0\\
                  &                   & 3 &51.8&57.8&49.7&57.9&59.3&53.4&60.2&60.8&56.5&59.6&60.9&57.0&60.1&61.2&57.6&60.2&60.9&57.3\\
                  &                   & 4 &53.4&57.7&49.4&58.0&59.7&54.4&59.9&60.7&57.0&59.6&60.7&57.2&59.7&60.9&57.3&60.5&61.4&58.4\\
                  \hline
\multirow{8}{*}{LSTM} & \multirow{4}{*}{RG} & 1 &62.3&62.6&62.0&62.8&63.0&62.3&63.0&63.1&62.4&62.0&62.1&61.6&62.6&62.8&62.2&62.2&62.4&61.7\\
                  &                   & 2 &62.5&62.5&61.8&61.8&61.7&61.2&61.4&61.4&60.6&60.5&60.7&59.8&60.3&60.4&59.7&60.5&60.6&59.9\\
                  &                   & 3 &60.7&60.8&60.1&61.3&61.2&60.7&60.7&60.6&59.7&60.0&60.0&59.3&60.2&60.1&59.5&60.5&60.3&59.7\\
                  &                   & 4 &60.6&60.5&59.9&60.7&60.7&59.9&59.4&59.2&58.4&58.7&58.6&57.6&59.3&59.2&58.6&59.1&59.1&58.4\\
                  \cline{2-21}
                  & \multirow{4}{*}{CG} & 1 &62.9&63.3&62.4&62.8&63.2&62.3&62.3&62.6&61.8&62.9&63.2&62.5&62.0&62.4&61.6&61.6&62.0&61.1\\
                  &                   & 2 &62.6&62.8&61.9&61.8&61.9&61.1&61.3&61.6&60.5&61.2&61.4&60.6&61.7&61.9&60.9&60.3&60.7&59.5\\
                  &                   & 3 &61.8&61.9&61.2&61.1&61.2&60.3&60.9&61.0&59.9&60.6&60.8&59.7&59.9&60.3&59.0&59.7&59.9&59.1\\
                  &                   & 4 &61.6&61.7&60.8&61.5&61.3&60.5&60.1&60.3&59.2&59.5&59.7&58.5&59.5&59.7&58.5&59.4&59.7&58.6\\
                  \hline
\multirow{8}{*}{BLSTM} & \multirow{4}{*}{RG} & 1 &62.8&62.9&62.4&62.2&62.5&61.9&62.3&62.4&61.9&61.9&62.0&61.4&62.3&62.5&61.6&62.8&63.0&62.5\\
                  &                   & 2 &62.6&62.5&62.0&62.0&62.1&61.6&62.1&62.1&61.6&61.1&61.1&60.6&61.3&61.4&60.7&60.8&61.0&60.3 \\
                  &                   & 3 &61.4&61.4&60.9&61.4&61.2&60.6&61.1&60.8&60.3&60.2&60.4&59.7&59.9&60.0&59.2&60.3&60.3&59.5 \\
                  &                   & 4 &60.0&60.0&59.4&61.1&61.0&60.5&60.5&60.5&60.1&60.2&60.3&59.6&60.2&60.3&59.6&60.0&60.1&59.2\\
                  \cline{2-21}
                  & \multirow{4}{*}{CG} & 1 &62.5&62.9&61.9&62.4&62.9&62.0&63.1&63.6&62.7&63.0&63.1&62.1&62.1&62.7&61.8&62.0&62.4&61.7\\
                  &                   & 2 &62.5&62.7&61.9&62.2&62.1&61.3&61.8&62.0&60.9&62.1&61.9&61.5&61.2&61.5&60.5&60.6&61.0&60.3\\
                  &                   & 3 &60.8&61.0&60.1&61.5&61.3&60.6&60.9&61.2&60.5&61.7&61.9&60.8&60.3&60.7&59.8&60.3&60.7&59.8\\
                  &                   & 4 &61.4&61.2&60.3&61.2&61.2&60.4&60.8&61.2&60.3&60.6&60.8&60.0&60.8&61.1&60.2&60.0&60.5&59.6\\
                  \hline
                  \hline
\end{tabular}
\begin{tablenotes}\footnotesize
\item[] {Sub. = subjects, RG = resting group, CG = comprehensive group, HT = hidden unit type, HL = number of hidden layers, HU = number of hidden units.}
\end{tablenotes}
\end{threeparttable}
\end{table*}

\section{Conclusion}
We present a novel method for automatic sleep stage classification for wearable devices using heart rate and wrist actigraphy. The method consists of two phases: the multi-level feature learning framework and the BLSTM-based RNN classifier. Unlike traditional approaches which focus on hand-engineered features, feature extraction in this work is designed to capture properties of raw sleep data and composition-based structural representation. RNN learns temporally sequential patterns of sleep. We have demonstrated the effectiveness of the proposed method. Our method is suitable for utilization in wearable devices. As little prior domain knowledge is used for the task, our method can be applied to other sleep-related issues, for example, sleep disorders diagnose. Future work will be conducted on improving automatic scoring performance.

\section*{Acknowledgment}
The authors would like to thank Microsoft Research Asia for providing the Microsoft Band 1 and the General Hospital of the Air Force, PLA, Beijing, China for scoring sleep data.




%

\bibliographystyle{IEEEtran}
\bibliography{manuscript}

%

%





\end{document}